\documentclass[letterpaper, 12pt]{article}
\usepackage{epsf}
\usepackage{times,bm,mathrsfs} 
\usepackage{amsmath}
\usepackage{amssymb}
\usepackage{amsfonts}
\usepackage{amsthm}
\usepackage{bbm}
\usepackage{color}
\usepackage{graphicx}
\usepackage{nccbbb}

\newcommand{\acal}{\mathcal{A}}

\newcommand{\ccal}{\mathcal{C}}
\newcommand{\dcal}{\mathcal{D}}

\newcommand{\ical}{\mathcal{I}}

\newcommand{\mcal}{\mathcal{M}}
\newcommand{\ncal}{\mathcal{N}}

\newcommand{\tcal}{\mathcal{T}}

\newcommand{\intgr}{\mathbb{Z}}
\newcommand{\nintgr}{\mathbb{N}}

\newcommand{\eps}{\varepsilon}

\newcommand{\ind}{\bbbone}

\newcommand{\bdm}{\begin{displaymath}}
\newcommand{\edm}{\end{displaymath}}
\newcommand{\bea}{\begin{eqnarray*}}
\newcommand{\eea}{\end{eqnarray*}}
\newcommand{\bean}{\begin{eqnarray}}
\newcommand{\eean}{\end{eqnarray}}

\newcommand{\prob}{\mathbb{P}}
\newcommand{\expec}{\mathbb{E}}

\newcommand{\poly}{\mathrm{poly}}

\newcommand{\hs}{\texttt{HS}}
\newcommand{\bs}{\texttt{BS}}
\newcommand{\sgn}{\mathrm{sgn}}

\newtheorem{theorem}{Theorem}

\newtheorem{corollary}{Corollary}

\newtheorem{lemma}{Lemma}

\author{{\bf S\'ebastien Roch} \\ 
Department of Statistics\\
University of California, Berkeley\\
Berkeley, CA 94720-3860\\
\texttt{\small sroch@stat.berkeley.edu}}

\begin{document}

\title{On Learning Thresholds of Parities and\\ 
Unions of Rectangles 
in Random Walk Models}

\maketitle

\begin{abstract}
In a recent breakthrough, [Bshouty et al., 2005] obtained
the first passive-lear\-ning algorithm for DNFs under
the uniform distribution. They showed that DNFs are learnable 
in the Random Walk and Noise Sensitivity models. 
We extend their results
in several directions. We first show that thresholds of
parities, a natural class encompassing DNFs, cannot be learned efficiently
in the Noise Sensitivity model 
using only statistical queries. In contrast, we show that
a cyclic version of the Random Walk model allows to learn efficiently 
polynomially weighted thresholds of parities.
We also extend the algorithm of Bshouty et al. to the case
of Unions of Rectangles, a natural generalization of DNFs to $\{0,\ldots,b-1\}^n$. 
\end{abstract}

\bigskip


\section{Introduction}

Learning Boolean formulae in Disjunctive Normal Form (DNF) 
has been a central problem in the computational
learning theory literature since Valiant's seminal paper
on PAC learning~\cite{Va84}. In~\cite{Ja94}, it was shown that DNFs
can be learned 
using membership queries, a form of active learning.
Jackson's algorithm, also known as Harmonic Sieve (\hs), 
uses a clever combination of two fundamental techniques in learning,
Harmonic Analysis and Boosting. 
The use of Harmonic Analysis in the study of Boolean functions was introduced
in~\cite{KKL88}. It was subsequently used as the basis of a learning
algorithm for $AC^0$ circuits in~\cite{LMN89}. The Harmonic Analysis 
used in the \hs\ algorithm
is based on a parity-finding algorithm of Goldreich and Levin~\cite{GL89}, which 
was first applied to a learning problem by Kushilevitz and Mansour~\cite{KM91}. 
Hypothesis boosting, a technique to reduce the classification error of a learning
algorithm,
was introduced by Schapire~\cite{Sc89}. The boosting
algorithm used by \hs\ is actually due to Freund~\cite{Fr90}.

In a recent breakthrough, Bshouty et al.~\cite{BM+03} obtained
the first passive learning algorithm for DNFs. Their algorithm
is based on a modification of \hs\ which focuses on low-degree
Fourier coefficients. That variant of \hs, called Bounded Sieve (\bs), 
was first obtained in~\cite{BF02}. 
In~\cite{BM+03}, \bs\ was used to learn
DNFs under the uniform distribution in two natural passive learning
models. The first one is the Random Walk model, where examples, instead
of being i.i.d., follow a random walk on the 
Boolean cube (see also~\cite{BFH94, Ga99} for related work). The second
model is the closely related Noise Sensitivity model, where this time examples
come in pairs, the second instance being a noisy version of the first one.
The results of~\cite{BM+03} are interesting in that they give
a learning algorithm for DNFs in a case where the observer has no control
over the examples provided. However the problem of learning DNFs under
the uniform distribution when examples are i.i.d. still remains open.
It is known that DNFs cannot be learned in the more restrictive
Statistical Query model (introduced in~\cite{Ke93}) where one can ask only about statistics
over random examples~\cite{BF+94}.

Jackson~\cite{Ja94} also showed that \hs\ applies to
thresholds of parities (TOP), a class that can express DNFs and decision trees
with only polynomial increase in size,
and extended his algorithm to the non-Boolean case of unions
of rectangles, a generalization of DNFs to $\{0,\ldots,b-1\}^n$
(where $b=O(1)$). 
Whether those classes of functions can be learned
in the Random Walk and Noise Sensitivity models was left open
by~\cite{BM+03}. Our contribution is threefold. We first show that
TOPs cannot be learned in the Noise Sensitivity model
using
statistical queries (SQs)\footnote{\cite{BM+03} uses only SQs.}. As far
as we know, this is the first example of a negative result for
``second-order'' statistical queries, i.e. queries on pairs
of examples. This does not rule out
the possibility of learning TOPs in the Random Walk model although
it provides evidence that the techniques of~\cite{BM+03} cannot be
easily extended to that case. On the other hand, we show that a simple
variant of the Random Walk model where the component updates follow a fixed
cycle allows to learn TOPs efficiently. This seems to be the
first not-too-contrived passive model in which TOPs are efficiently
learnable with respect to the uniform distribution. Actually,
one can perform the Harmonic Sieve in this Cyclic Random Walk model, and we also show
that this model is strictly weaker than the active setting under
a standard cryptographic assumption. Finally we extend 
the techniques
of~\cite{BF02} and \cite{BM+03} to 
the non-Boolean domain $\{0,\ldots,b-1\}^n$ and use this
to learn unions of rectangles in the Noise Sensitivity and Random Walk models. 
This last result turns out to be rather straightforward once
the proper analogues to the Boolean case are found.

In Section~\ref{sec:prelim}, we introduce the learning models and give a brief review
of Fourier analysis. The negative result for learning TOPs is derived in Section~\ref{sec:neg}.
The learning algorithms for TOPs and Unions of Rectangles are presented in
Sections \ref{sec:cyc} and \ref{sec:ubox} respectively.

\section{Preliminaries}\label{sec:prelim}

We briefly review the learning models we will use and some basic facts about Fourier analysis.
For more details see e.g. \cite{KV94} and \cite{Ma94}.

\subsection{Learning Models}

Let $b\in\nintgr$ be a nonzero constant and let $[b] = \{0,\ldots,b-1\}$.
Often we will take $b=2$.
Consider a function $f:[b]^n \to \{1,-1\}$, that we will call
the {\em target} function. Think of $f$ as 
partitioning $[b]^n$ into positive and negative examples. Denote
by $U$ the uniform distribution over $[b]^n$. 
The goal of the different learning problems we will consider
is generally to find for $\eps > 0$ an $\eps$-approximator $h$ to $f$
under the uniform distribution,
i.e. a function $h$ such that\footnote{
For convenience, we will drop the notation $x\sim U$ from 
probabilities and expectations when it is clear that $x$ is uniform. 
}
\bea
\prob_{x\sim U}[h(x) \neq f(x)] \leq \eps.
\eea
To achieve this, the learner is given access to limited information
which can take different forms. 

The {\bf Membership Query} (MQ)
model allows to ask for the value of $f$ at any point $x$ of our
choosing. The {\bf Uniform Query} (UQ) model on the other hand works as follows:
at any time the learner can ask for an example from $f$
and is provided with a pair
$\langle x, f(x) \rangle$ where $x \sim U$; all examples are independent. 
This type of model is called passive---contrary to the MQ
model which is called active---because the learner has
no influence over the example provided to him.

In~\cite{BM+03}, two variants of this model were considered.
In the {\bf Random Walk} (RW) model, one is given access to random
examples $\langle x, f(x) \rangle$ where the successive values of
$x$ follow a random walk on $[b]^n$. Many choices of walks
are possible here. We will restrict ourselves to the case
where at each step, one component of $x$, say $x_i$, is
picked uniformly at random and a new value $y$ for $x_i$
is picked uniformly at random over $[b]$
(the first example is uniform over $[b]^n$). A related model
is the {\bf Noise Sensitivity} (NS) model. Here a parameter $\rho\in[0,1]$
is fixed and when an example is asked, one gets $\langle x,y,f(x),f(y),S \rangle$
where $x\sim U$ and $y\equiv\ncal_{\rho}(x)$ is a noisy version of $x$ defined as follows:
for each component of $x$ independently with probability $1-\rho$
a new uniform value over $[b]$ is drawn for this component
(we call this operation {\em updating} and we call $1 - \rho$ the
{\em attribute noise rate}), otherwise
the component remains the same\footnote{Note that a component
is allowed to remain the same even if it is updated.};
$S$ is the set of updated components. We will consider one more variant
of these passive models. In the {\bf Cyclic Random Walk} (CRW) model,
the successive examples $x$ follow a random walk where at each step, instead of picking
a uniformly random component to update, there is a fixed
cycle $(i_1,\ldots,i_n)$ running through all of $\{1,\ldots,n\}$ 
and components are updated in that order (the first example is uniform over $[b]^n$).
In all the previous models except MQ,
examples are drawn randomly and we therefore allow the learning
algorithm to err with probability $1-\delta$ for some $\delta$.

The UQ and NS models also have a {\bf Statistical Query} (SQ)
variant. Here, one does not have access to actual examples. Instead
in the case of UQ for instance one can choose a polynomial-time computable function 
$\Gamma : [b]^n \times \{1,-1\} \to \{1,-1\}$ and a tolerance
$\tau \in [0,1]$ which is required to be at least inverse polynomially
large and the UQ-SQ oracle returns a number $\gamma$ such
that
\bea
|\expec[\Gamma(x, f(x))] - \gamma| \leq \tau.
\eea
Therefore, the learner can ask only about statistics over
random examples.
This can be simulated in polynomial time under the UQ model
using empirical averaging. But the UQ model is strictly more
powerful than UQ-SQ~\cite{Ke93}. In the case of NS,
the function $\Gamma$ is allowed to depend on $x,y,f(x),f(y)$. 
This is called a {\em second-order} statistical query.

We will work with three classes of functions. First,
we consider Boolean formulae in Disjunctive Normal
Form (DNF), in which case $b=2$. A natural
generalization of DNFs to $b>2$ was given in~\cite{Ja94}:
for each $1 \leq i \leq n$, choose two values 
$0 \leq l_i \leq u_i \leq b-1$, and consider the
{\em rectangle}
\bea
[l,u] = \{x\in[b]^n\ :\ l_i \leq x_i \leq u_i,\ \forall i\}.
\eea
An instance of UBOX is a union of rectangles. Note that
in the Boolean case, a DNF can be seen as a union of subcubes
of $[2]^n$. The class of thresholds of parities (TOP) applies
only to $b=2$. A TOP is a function of the form
\bea
f(x) = \sgn\left(\sum_{m=1}^M w_m (-1)^{\sum_{i} a_i^{(m)} x_i}\right),
\eea
for $M$ vectors $a^{(m)} \in [2]^n$ and weights $w_m \in \intgr$. 
It is assumed that the weight sum $\sum_{m=1}^M |w_m|$ 
is of size polynomial in $n$.

We will be interested in learning function classes under the uniform
distribution. For any model $\mcal$, any function class $\ccal$ and
any $\delta, \eps > 0$, we say that $\ccal$ is $(\delta, \eps)$-learnable
in $\mcal$ if there is an algorithm $\acal$ such
that for any function $f\in\ccal$ with probability at least $1- \delta$,
$\acal$ finds an $\eps$-approximator to $f$ in time polynomial in the
description size of $f$. We say that $\ccal$ can be weakly learned
under $\mcal$ if there is $\delta > 0$ and $\eps$ of the form
$\frac{1}{2} - \frac{1}{\poly(n)}$ such that $\ccal$ can
be $(\delta,\eps)$-learned in $\mcal$.

\subsection{Fourier Analysis}

The complex-valued\footnote{In the Boolean case, we actually
consider only real-valued functions.} functions on $[b]^n$ form a linear space
where a natural inner product is given by
\bea
\langle f, g\rangle = \frac{1}{b^n} \sum_{x} f(x)g^*(x) = \expec[f(x)g^*(x)],
\eea
where ${}^*$ denotes complex conjugation. The set of all generalized 
parities (parities for short)
\bea
\chi_a(x) = \omega_b^{\sum_i a_i x_i},
\eea
where $a \in [b]^n$ and $\omega_b = e^{2\pi i/b}$ form an orthonormal
basis and any function can be written as a linear combination
\bean\label{eq:comb}
f(x) = \sum_{a\in[b]^n} \hat f(a) \chi_a^*(x),
\eean
where the Fourier coefficient $\hat f(a)$ is $\expec[f(x)\chi_a(x)]$. 
A useful result is Parseval's identity
\bea
\expec[|f(x)|^2] = \sum_{a\in [b]^n} |\hat f(a)|^2.
\eea

In learning problems, Fourier-based algorithms usually estimate
some of the Fourier coefficients and build an approximation
to $f$ in the form of a linear combination as in (\ref{eq:comb}) (and then take
the sign or something slightly more complicated in the case $b>2$). There are
two main cases where this technique tends to work. In the ``low-degree'' case,
most of (or at least a non-negligible part of) the Fourier mass is concentrated
on low-degree terms, i.e. terms $\hat f(a)$ where $a$ has few non-zero components.
Then one can estimate all low-degree terms, which can lead to a
subexponential algorithm. This is the idea behind the algorithm
for learning $AC^0$ circuits in~\cite{LMN89}.
In the ``sparse'' case, most of the mass is
concentrated on a few terms. Then one needs to find a way
to determine which terms should be estimated. This is the idea behind the algorithm
for learning decision trees in~\cite{KM91}.

Because one often needs to estimate expectations, e.g. Fourier
coefficients, using
empirical averages, it is customary at this point to recall
Hoeffding's lemma.
\begin{lemma}[Hoeffding] Let $X_i$ be independent
random variables all with mean $\mu$ such that
for all $i$, $c \leq X_i \leq d$. Then
for any $\lambda > 0$,
\bea
\prob\left[\left|\frac{1}{m} \sum_{i=1}^m X_i - \mu\right| \geq \lambda \right]
\leq 2 e^{-2\lambda^2 m/(d-c)^2}.
\eea
\end{lemma}

\section{Negative Result for TOPs Learning}\label{sec:neg}

For this section, we fix $b=2$. 
As demonstrated in~\cite{Ke93} and \cite{BF+94}, a nice feature of the
SQ model is that it allows a complete unconditional characterization
of what is learnable under this model. We prove in this section that
parities cannot be 
weakly learned in the Noise Sensitivity model
with attribute noise rate at least $\frac{\omega(\log n)}{n}$ (this includes
the constant noise rate case used in~\cite{BM+03}). This implies in turn that TOPs cannot be
weakly learned in this model. Our lower bound on the noise rate
is tight for this impossibility result. Indeed, it is easy to see that for an attribute
noise rate of $\frac{O(\log n)}{n}$, one can actually learn parities. This
follows from the fact that at such a rate, there is a non-negligible
probability of witnessing an example $\langle x,y,f(x),f(y),S\rangle$ with
exactly one bit flip from $x$ to $y$, which allows to decide whether the updated variable
is contained or not in the parity. One can then repeat for all
variables (this can be turned into a statistical query test). 
In this section, $y=\ncal_\rho(x)$ with $x$ uniform unless
stated otherwise.

We follow a proof of~\cite{BF02}. The main difference
is that we need to deal with second-order queries.
\begin{lemma}\label{lemma:queries}
Any SQ, $\Gamma(x,y,f(x),f(y))$, in the NS model can be replaced
by simple expectations, $1$st-order queries of the form
$\expec[g(x)f(x)]$ (where $x$ is uniform), and 
$2$nd-order queries of the form $\expec[h(x,y)f(x)f(y)]$ where
$f$ is the target function (this actually applies
to any second-order SQ model). Moreover, we can assume
$|g(x)| \leq 1$ and $|h(x,y)| = 1$ for all $x,y \in [2]^n$.
\end{lemma}
\begin{proof} 
Say we are trying to learn the function $f$. Because $f$ takes only
values $-1$ and $+1$, we have
{\small
\bea
\expec[\Gamma(x,y,f(x),f(y))]
&=& \expec\left[\sum_{i,j=+1,-1}\Gamma(x,y,i,j)\left(\frac{1 + if(x)}{2}\right)
\left(\frac{1 + j f(y)}{2}\right) \right]\\
&=& \frac{1}{4}\sum_{i,j=+1,-1} \Big(\expec[\Gamma(x,y,i,j)]
+ i \expec_x[f(x)\expec_y[\Gamma(x,y,i,j)]]\\
&& + j \expec_y[f(y)\expec_x[\Gamma(x,y,i,j)]]
+ ij \expec[f(x)f(y)\Gamma(x,y,i,j)]\Big).
\eea 
}
\end{proof}

Note that the $1$st-order queries may not be computable in polynomial time
because the averages over $x$, $y$ are exponential sums (although they might
be estimated in polynomial time). But this
is not a problem because what we will show is that, no matter what the complexity
of the queries is, the {\em number} of queries has to be superpolynomial.
Note also that the simple expectations do not require the oracle (assuming
the distribution of $x,y$ is known, as is the case in the NS model).
So we ignore them below. Finally, note that in the NS-SQ model, expectations
are unchanged if the roles of $x$ and $y$ are reversed.

Following Lemma~\ref{lemma:queries}, we can think of a weakly learning algorithm as
making a polynomial number of $1$st and $2$nd-order queries. 
Denote by $s$ the size of the target function. Say the algorithm $\acal$
makes $p(n,s)$ queries with tolerance $1/r(n,s)$ and outputs an 
$(\frac{1}{2} - \frac{1}{q(n,s)})$-approximator, where the queries 
are a collection of functions 
$\{(g^{n,s}_i(x), h^{n,s}_i(x,y))\}_{i=1}^{p(n,s)}$
over $x,y \in [2]^n$ with $|g^{n,s}_i(x)| \leq 1$ and
$|h^{n,s}_i(x,y)| = 1$ for all $x,y \in [2]^n$. 
We now characterize weakly learnable classes in NS-SQ
(the characterization actually applies to any second-order SQ model).
For this proof, we assume that the confidence parameter $\delta= 0$
(but see the remark after the proof).
\begin{lemma}\label{lemma:charac}
Let $r'(n,s) = \max\{2r(n,s), q(n,s)\}$.
Denote by $\ccal^{n,s}$ the class of functions in $\ccal$ restricted to instances
of $n$ variables and size at most $s$.
If $\ccal$ is weakly learnable under NS-SQ (using an algorithm
with parameters described above), then
there exists a collection $\{V_{n,s}\}_{n,s\geq 1}$ with $V_{n,s}$ of the form
\begin{equation*}
\{(k^{n,s}_i(x), l^{n,s}_i(x,y))\}_{i=1}^{p'(n,s)}
\end{equation*}
with $|k^{n,s}_i(x)| \leq 1$ and $|l^{n,s}_i(x,y)|=1$ for all $x,y \in [2]^n$,
and $p'(n,s) \leq p(n,s) + 1$ such
that,
\bean\label{eq:witness}
\forall f\in\ccal^{n,s},\ \exists i,\ |\expec_x[f(x)k^{n,s}_i(x)]| + 
|\expec[f(x)f(y)l^{n,s}_i(x,y)]| \geq \frac{2}{r'(n,s)}.
\eean
\end{lemma}
\begin{proof}
We start with $V_{n,s} = \emptyset$.
We simply simulate the weak learning algorithm $\acal$ with an oracle that 
returns the value $0$ to each query. Every time $\acal$ makes a query, we add
that query to $V_{n,s}$. At one point $\acal$ stops and returns the hypothesis $\sigma$.
We add $(\sigma,1)$ to $V_{n,s}$. It is clear that $p'(n,s) \leq p(n,s) + 1$. Assume
that (\ref{eq:witness}) is not satisfied. Then there is a function
$f$ such that 
\bea
|\expec_x[f(x)k^{n,s}_i(x)]| < \frac{2}{r'(n,s)},\qquad 
|\expec[f(x)f(y)l^{n,s}_i(x,y)]| < \frac{2}{r'(n,s)},
\eea
for all $i$. Therefore, in our simulation, the zeros we gave as answers to the queries
were valid answers (i.e. within the tolerance $\frac{1}{r(n,s)}$) and therefore because $\acal$ returns
a weak approximator, it has to be the case that $\sigma$ is a 
$(\frac{1}{2} - \frac{1}{q(n,s)})$-approximator. This implies that
\bea
|\expec[f(x)\sigma(x)]| + |\expec[f(x)f(y)1]| \geq |\expec[f(x)\sigma(x)]| 
\geq \frac{2}{q(n,s)} \geq \frac{2}{r'(n,s)},
\eea
a contradiction.
\end{proof}

As noted in~\cite{BF02}, because the previous proof does not rely on the uniformity
of the learning algorithm and because $\mathrm{BPP} \subseteq \mathrm{P}/\poly$, the proof
also applies to randomized algorithms.

\begin{theorem}\label{thm:neg}
The class of parity functions cannot be weakly learned in NS-SQ with attribute noise
rate $\frac{\omega(\log n)}{n}$. 
\end{theorem}
\begin{proof}
Because the size of the function is bounded by a polynomial in $n$, we drop
$s$ from the previous notations. Suppose to the contrary that there is an algorithm $\acal$ with parameters
as described above that weakly learns parities. By Lemma~\ref{lemma:charac}, we have
for all $a\in[2]^n$
\bea
\sum_{i=1}^{p'(n)} \Big(\expec^2_x[\chi_a(x)k^n_i(x)] + \expec^2[\chi_a(x)\chi_a(y)l^n_i(x,y)]\Big) 
\geq \frac{2}{(r'(n))^2}.
\eea
Taking expectation over uniform
$a\in [2]^n$, this is
\bea
\sum_{i=1}^{p'(n)} \expec_a[\expec^2_x[\chi_a(x)k^n_i(x)]] + \sum_{i=1}^{p'(n)} \expec_a[\expec^2[\chi_a(x)\chi_a(y)l^n_i(x,y)]] 
\geq \frac{2}{(r'(n))^2}.
\eea
Then either 
\bean\label{eq:sumk}
\sum_{i=1}^{p'(n)} \expec_a[\expec^2_x[\chi_a(x)k^n_i(x)]] \geq \frac{1}{(r'(n))^2},
\eean
or
\bean\label{eq:suml}
\sum_{i=1}^{p'(n)} \expec_a[\expec^2[\chi_a(x)\chi_a(y)l^n_i(x,y)]]\geq \frac{1}{(r'(n))^2}.
\eean
In case (\ref{eq:sumk}), we get a contradiction by following the same
steps as in~\cite[Theorem 34]{BF02}, which we do not repeat here
(their $k$ becomes $n$ and their $\rho$ becomes $\frac{1}{2}$). 
The attribute noise rate does not play a role in that case.
Below, we
derive a contradiction out of (\ref{eq:suml}), which follows a similar
argument. 

From (\ref{eq:suml}) there is an $i$ such that
\bean\label{eq:firsti}
\ical = \expec_a[\expec^2[l^n_i(x,y)\chi_a(x)\chi_a(y)]] \geq \frac{1}{p'(n)(r'(n))^2}.
\eean
Taking $(u,v)$ to be an independent copy of $(x,y)$, we also have
\bea
\ical &=& \expec_a[\expec[l^n_i(x,y)\chi_a(x)\chi_a(y)]
\expec[l^n_i(u,v)\chi_a(u)\chi_a(v)]]\\
&=& \expec_{(x,y)}[\expec_{(u,v)}[l_i^n(x,y)l_i^n(u,v) \expec_a[\chi_a
(x\oplus y\oplus u\oplus v)]]],
\eea
where $\oplus$ is the parity operator.
Denote $\gamma_1 = x\oplus u$ and $\gamma_2 = y\oplus v$. Recall that
$|l_i^n(x,y)| = 1$  for all $x,y \in [2]^n$. Then
\bea
|\ical| 
&\leq& \expec_{\gamma_1,\gamma_2}[|\expec_a[\chi_a(\gamma_1 \oplus \gamma_2)]|]\\
&=& \expec_{\gamma_1,\gamma_2}[|\expec_a[\chi_{\gamma_1 \oplus \gamma_2}(a)]|]\\
&=& \expec_{\gamma_1}[\prob_{\gamma_2}[ \gamma_1 = \gamma_2]]\\
&=& \expec_{\gamma_1}\left[\left(\rho^2 + \frac{1}{2}(1 - \rho^2)\right)^n\right]\\
&=& \left(1 - (1 - \rho) + \frac{1}{2}(1-\rho)^2\right)^n.
\eea
This last term is the inverse of a superpolynomial if 
$(1 - \rho) = \frac{\omega(\log n)}{n}$, which contradicts (\ref{eq:firsti}).
\end{proof}

In the case of constant attribute noise rate, the proof actually implies that
even the parities over the first $\omega(\log n)$ variables cannot be
weakly learned. 

\section{Harmonic Sieve in Cyclic Random Walk Model}\label{sec:cyc}

In this section, we show that \hs\ can be performed efficiently in the CRW model.
We also prove that CRW is strictly weaker than MQ under a standard
cryptographic assumption.
\begin{theorem}\label{thm:hscrw}
The algorithm \hs\ can be performed in the CRW model with a polynomial increase
in time (and an arbitrarily small probability of error).
\end{theorem}
As an immediate corollary we get the following.
\begin{corollary}
For any $\delta, \eps > 0$, DNFs, TOPs and UBOXs are
$(\delta,\eps)$-learnable in the CRW model.
\end{corollary}
The proof of Theorem~\ref{thm:hscrw} follows.
\begin{proof} 
We only need to check that we can estimate the sums of
squares of Fourier coefficients appearing in the
Goldreich-Levin algorithm. Without loss of generality,
we can rename all components of $x$ so that the
components are updated in the order $(n, n-1,\ldots, 1)$.
For $1 \leq k \leq n$ and $a\in [b]^k$, let
\bea
C_{a,k} = \{\hat f(ad)\ :\ d\in[b]^{n-k}\},
\eea
where $ad$ is the concatenation of $a$ and $d$. Then Jackson~\cite{Ja94}
showed that it is enough to estimate within 
inverse polynomial additive tolerance the sum of the squares
of terms in $C_{a,k}$ which he also shows to be equal to
\bea
L^2_2(C_{a,k}) = \sum_{d\in[b]^{n-k}} \hat f^2(ad) = 
\expec[\mathrm{Re}(f^*(yx) f(zx) \chi_a(y-z))],
\eea
where $x \in [b]^{n-k}$, $y\in [b]^k$ and $z\in [b]^k$
are independent uniform, and $y-z$ is taken to be the difference
in $\intgr_b^k$. In the CRW model, this estimation can be achieved through
the following simulation. Make $n$ queries to obtain
a uniform instance. Then make $n-k$ queries to update
the last $n-k$ bits and get $yx$ and $f(yx)$. Then make
$k$ more queries to update the first $k$ bits and
get $zx$ and $f(zx)$. It is clear that
$x,y,z$ are as required above. From this,
compute $\mathrm{Re}(f^*(yx) f(zx) \chi_a(y-z))$. Repeat
sufficiently (polynomially) many times and apply
Hoeffding's lemma. This takes $2n$ times as many
queries as in the MQ model. The rest of the \hs\ algorithm
applies without change. Note, in particular, that 
the boosting part does not require membership queries
(see also~\cite[Theorem 21]{BF02}). Note also that we didn't assume
that $f$ is Boolean above.
\end{proof}

\begin{theorem}\label{thm:crwmq}
If one-way functions exist, the CRW model is strictly
weaker than the MQ model.
\end{theorem}
\begin{proof}
We proceed as in~\cite[Proposition 2]{BM+03}. If one-way
functions exist then there exists a pseudorandom
function family $\{f_s : [2]^n \to \{1, -1\}\}_{s\in\{1, -1\}^n}$
\cite{HI+90}.
Consider the function $g_s$ which is equal to $f_s$ except
on inputs of the form $e_i$ (i.e. the vector with $0$'s everywhere except 
on component $i$ where it is $1$) where the function
is defined as $s_i$. Then using membership queries,
one can learn $s$ from queries to $g_s$ and therefore one can learn
$g_s$. On the other hand, in the CRW model, with probability
$1 - 2^{-\Omega(n)}$, one never sees instances $e_i$'s. Therefore
if it were possible to learn $g_s$ in this model, this would be
essentially equivalent to efficiently learning $f_s$ in the MQ model 
(by simulation of the conditioned walk)
which leads to a contradiction.
\end{proof}

\section{Learning Unions of Rectangles}\label{sec:ubox}

The purpose of this section is to extend the DNF learning
algorithm of~\cite{BM+03} in the Noise Sensitivity model
to the $[b]^n$ setting. The learning algorithm of~\cite{BM+03}
proceeds in a fashion similar to that of~\cite{Ja94} except
that it uses {\em weighted} sums of squared Fourier coefficients
(related to the so-called Bonami-Beckner operator)
and considers only $O(\log n)$-degree terms.
Therefore the main task in extending this algorithm
to UBOXs is to define an appropriate substitute for the
Bonami-Beckner operator and show that low-degree terms
are also sufficient in this case. The latter was proved
by Jackson~\cite[Corollary 17]{Ja94}. We tackle
the former problem in the following theorem. 
\begin{theorem}\label{thm:ubox} 
For any $\delta, \eps > 0$, the class of UBOXs is $(\delta,\eps)$-learnable in the
Noise Sensitivity model, and therefore in the
Random Walk model as well.
\end{theorem}
\begin{proof}
We seek to generalize the weighted sum of squared coefficients used
in~\cite{BM+03}. A requirement is that it must be possible to estimate
the partial sums corresponding to fixing $O(\log n)$ components in
the Noise Sensitivity model. A natural choice seems to be
\bea
(T_\rho f)(x) = \expec_{y = \ncal_{\rho}(x)}[f(y)],
\eea
where recall that $\ncal_{\rho}(x)$ is a noisy version
of $x$ where each component is updated independently with probability
$1 - \rho$. Here $\rho$ is a fixed constant. Because the operator $T_\rho$ is linear, it suffices
to compute its action on the basis functions. 
Denote by $|S|$ the cardinality of $S \subseteq \{1,\ldots,n\}$ and by 
$|a|$ the number of nonzero components of $a \in [b]^n$. For
a vector $x$ and a set $S$, we note $x_S$ the vector $x$ restricted
to components in $S$, and $0_{S^c}x_S$ signifies the vector
which has $0$'s on components in $S^c$ and is equal to $x$
on components in $S$.
For any $a\in[b]^n$, we have
\bea
\expec_{y = \ncal_{\rho}(x)}[\chi_a(y)]
&=& \frac{1}{b^n}\sum_{z \in [b]^n} \sum_{m=0}^n \sum_{S:|S| = m} 
(1 - \rho)^m \rho^{n-m} \chi_a(x + 0_{S^c}z_S)\\
&=& \chi_a(x) \sum_{m=0}^n \sum_{S:|S| = m} 
(1 - \rho)^m \rho^{n-m} \frac{1}{b^n}\sum_{z \in [b]^n} \chi_{a_S}(z_S)\\
&=& \chi_a(x) \sum_{m=0}^n \sum_{S:|S| = m} 
(1 - \rho)^m \rho^{n-m} \ind\{|a_S| = 0\}\\
&=& \chi_a(x) \rho^{|a|} \sum_{m=0}^{n - |a|} \binom{n - |a|}{m} (1 - \rho)^m
\rho^{n - |a| - m}\\
&=&  \rho^{|a|} \chi_a(x).
\eea
Therefore,
\bea
(T_\rho f)(x) = \sum_{a \in [b]^n} \rho^{|a|} \hat f(a) \chi^*_a(x).
\eea
This kind of operator has been used before. See e.g.~\cite{Ja97}.

We are interested in partial sums of the form
\bea
\tcal(I) = \sum_{a:|a_I|=|I|} \rho^{|a|} |\hat f(a)|^2,
\eea
where $I \subseteq \{1, \ldots, n\}$. Indeed, those allow to perform
the breadth-first search algorithm in \cite[Theorem 7]{BM+03}. Note first
that we get a similar upper bound on the weighted Fourier mass of a fixed level
of the BFS tree
\bea
\sum_{I:|I|=j}\tcal(I)
&=& \sum_{I:|I|=j} \sum_{a:|a_I|=|I|} \rho^{|a|} |\hat f(a)|^2\\
&=& \sum_{a:|a|\geq j} \binom{|a|}{j} \rho^{|a|} |\hat f(a)|^2\\
&\leq& \sum_{a:|a|\geq j} |\hat f(a)|^2 \sum_{t=j}^{+\infty} \binom{t}{j} \rho^t\\
&\leq& \expec[|f(x)|^2] \rho^{-1} \left(\frac{\rho}{1 - \rho}\right)^{j+1}\\
&\leq& \max_x\{|f(x)|^2\} \rho^j (1 - \rho)^{-j-1},
\eea
where we have used Parseval's identity.
The rest of the proof of~\cite[Theorem 7]{BM+03} goes without change.
The only difference is that now to every $I \subseteq \{1,\ldots,n\}$
corresponds $(b-1)^{|I|}$ vectors $a\in[b]^n$ with $|a_I| = |I|$
and $|a_{I^c}| = 0$. But we can afford to estimate all of them because
$|I| = O(\log n)$. Therefore we can find all inversely polynomial
coefficients of order $O(\log n)$. Also, we need to check that any UBOX
has at least one inversely polynomial coefficient of order $O(\log n)$
and that boosting
is possible. This is
done in~\cite[Section 6]{Ja94}.
The only point to note is that in the proofs of~\cite[Fact 14, Corollary 17]{Ja94},
one can choose the parity $\chi_a$ to have all its components 0 outside the variables
included in the $O(\log n)$-rectangle used in the proof (see also~\cite[Lemma 18]{BF02}). 

It only remains to show that the $\tcal(I)$'s can be estimated in the
Noise Sensitivity model. As in~\cite{BM+03}, we consider the distribution
$\dcal_\rho^{(I)}$ over pairs $(x,y) \in [b]^n\times[b]^n$ which is
$(x, \ncal_\rho(x))$ conditioned on the event that at least all components
in $I$ are updated. This can be simulated in the Noise Sensitivity model
by simply picking examples $\langle x,y,f(x),f(y),S\rangle$ until one gets that 
$I \subseteq S$ (which takes polynomial time if $|I| = O(\log n)$). 
Then note that
{\small\bea
\tcal'(I)
&\equiv& \expec_{\dcal_\rho^{(I)}}[f(x)f(y)]\\
&=& \expec_{\dcal_\rho^{(I)}}\left[\sum_{c,d}\hat f(c)\hat f(d) \chi^*_c(x)\chi^*_d(y)\right]\\
&=& \frac{1}{b^{2n}} \sum_{x,z} \sum_{m = 0}^{n - |I|} \sum_{S:|S| = m + |I|}\sum_{c,d}
(1 - \rho)^m \rho^{n- |I| - m} \hat f(c)\hat f(d) \chi^*_c(x)\chi^*_d(x + 0_{S^c}z_S)\\
&=& \frac{1}{b^{2n}} \sum_{x,z} \sum_{m = 0}^{n - |I|} \sum_{S:|S| = m + |I|}\sum_{c,d}
(1 - \rho)^m \rho^{n- |I| - m} \hat f(c)\hat f(d) \chi^*_{c+d}(x)\chi^*_{d_S}(z_S)\\
&=& \sum_{m = 0}^{n - |I|} \sum_{S:|S| = m + |I|}\sum_{c,d}
(1 - \rho)^m \rho^{n- |I| - m} \hat f(c)\hat f(d) \ind\{c+d = 0\ \mathrm{mod}\ b\}
\ind\{|d_S| = 0\}\\
&=& \sum_{c} \sum_{m = 0}^{n - |I|} \sum_{S:|S| = m + |I|} (1 - \rho)^m \rho^{n- |I| - m}
|\hat f(c)|^2 \ind\{|c_S| = 0\}\\
&=& \sum_{c:|c_I| = 0} \rho^{|c|} |\hat f(c)|^2 \sum_{m = 0}^{n - |I| - |c|} 
\binom{n - |I| - |c|}{m}(1 - \rho)^m \rho^{n - |I| - m - |c|} \\
&=&  \sum_{c:|c_I| = 0} \rho^{|c|} |\hat f(c)|^2,
\eea}
\!\!where we have used that if $f$ is real and $c+d = 0\mod b$, then
\begin{eqnarray*}
\hat f(d) = \left(\hat f(c)\right)^*.
\end{eqnarray*}
Denote $\tcal''(I) = \tcal'(\emptyset) - \tcal'(I)$. This is
\bea
\tcal''(I) = \sum_{c:|c_I| > 0} \rho^{|c|} |\hat f(c)|^2.
\eea
We want to estimate $\tcal(I)$ which consists of a sum over
$\{a:|a_I|=|I|\}$. We now know how to estimate
the same sum over $\{a : |a_J| > 0\}$ for any $J$.
Noting that $\{a : |a_J| > 0\}$ is made precisely of all 
$\{a : |a_K|=|K|\}$ with $K\subseteq J$, it is easy to see
that $\tcal(I)$ can be estimated through the $\tcal''(J)$'s
for $J\subseteq I$ by inclusion-exclusion. Since there are only $2^{|I|}$
such $J$'s and $|I| = O(\log n)$, this can be done in polynomial time.
The rest of the argument is as in~\cite[Theorem 11]{BM+03}.
\end{proof}

\section*{Ackowledgements}

I thank Elchanan Mossel for comments and suggestions. I
gratefully acknowledge the financial support of NSERC (Canada)
and FQRNT (Quebec, Canada).

\end{document}